\documentclass{llncs}

\usepackage{graphicx} 
\usepackage{color}
\usepackage{coloring}

\usepackage{amsmath,amssymb,stmaryrd,textcomp}
\usepackage{proof}

\title{First Experiments with a  \\ Flexible Infrastructure for Normative Reasoning}
\author{Christoph Benzm\"uller \and Xavier Parent}
\institute{Computer Science and Communications, University of Luxembourg, Luxembourg} 

\begin{document}
\maketitle

\begin{abstract} 
A flexible infrastructure for normative reasoning is outlined. A small-scale demonstrator version of the envisioned system has been implemented in the proof assistant Isabelle/HOL by utilising the first authors universal logical reasoning approach based on shallow semantical embeddings in meta-logic HOL.
The need for such a flexible reasoning infrastructure is motivated and illustrated with a contrary-to-duty example scenario selected from the General Data Protection Regulation.  
\end{abstract}

\section{Introduction} 
We argue for the development of a flexible deontic logic reasoning infrastructure. This infrastructure shall support formalisation experiments in legal and ethical reasoning. Since the quest for the most suitable logical formalisms in this area is not yet settled, our infrastructure offers a range of deontic logic alternatives to be used and assessed in different application contexts. Our infrastructure is based on the first authors approach \cite{C66,R59} to universal logical reasoning via shallow semantical embeddings in meta logic HOL (classical higher-order logic). This approach enables the reuse of state-of-the-art theorem proving technology for the flexible mechanisation and automation of a range of non-classical logics.  The idea is to adapt this framework for various deontic logics, and to subsequently assess these logics in respective case studies. We illustrate the idea by
 first presenting embeddings of two alternative deontic logics in HOL (Section \ref{sec:embeddings}). These two alternative logics are then applied and assessed (in Section \ref{sec:experiments})
with an exemplary contrary-to-duty scenario we identified in the context of the General Data Protection Regulation (GDPR, Regulation EU
2016/679).

\section{Embedding Deontic Logics in HOL} \label{sec:embeddings}
Two different deontic logics are considered in this section: standard deontic logic (SDL) \cite{sep-logic-deontic} and the dyadic deontic logic (DDL) of Carmo and Jones \cite{CJ13,Carmo2002}. 
Both logics have been implemented in the proof assistant Isabelle/HOL \cite{Isabelle} by utilising the semantical embedding approach. The faithfulness of these embeddings in  HOL has been studied in previous work \cite{J23,R63}. The encoding of these logics could alternatively be carried out in any other theorem proving environment that entails classical higher-order logic HOL. For example, their encoding could well be carried out in the TPTP THF syntax \cite{J22} to enable the direct application of TPTP THF compliant higher-order automated theorem provers such as Satallax \cite{Satallax}, LEO-II \cite{J30} and Leo-III \cite{C70}, and the (counter-)model finder Nitpick \cite{BN10}. 
An example of an embedding of second-order modal logic KB (with constant domain quantifiers) in TPTP THF syntax is presented in \cite{W55}.
 
\begin{figure}[tp] \centering
\includegraphics[width=\textwidth,height=.65\textheight]{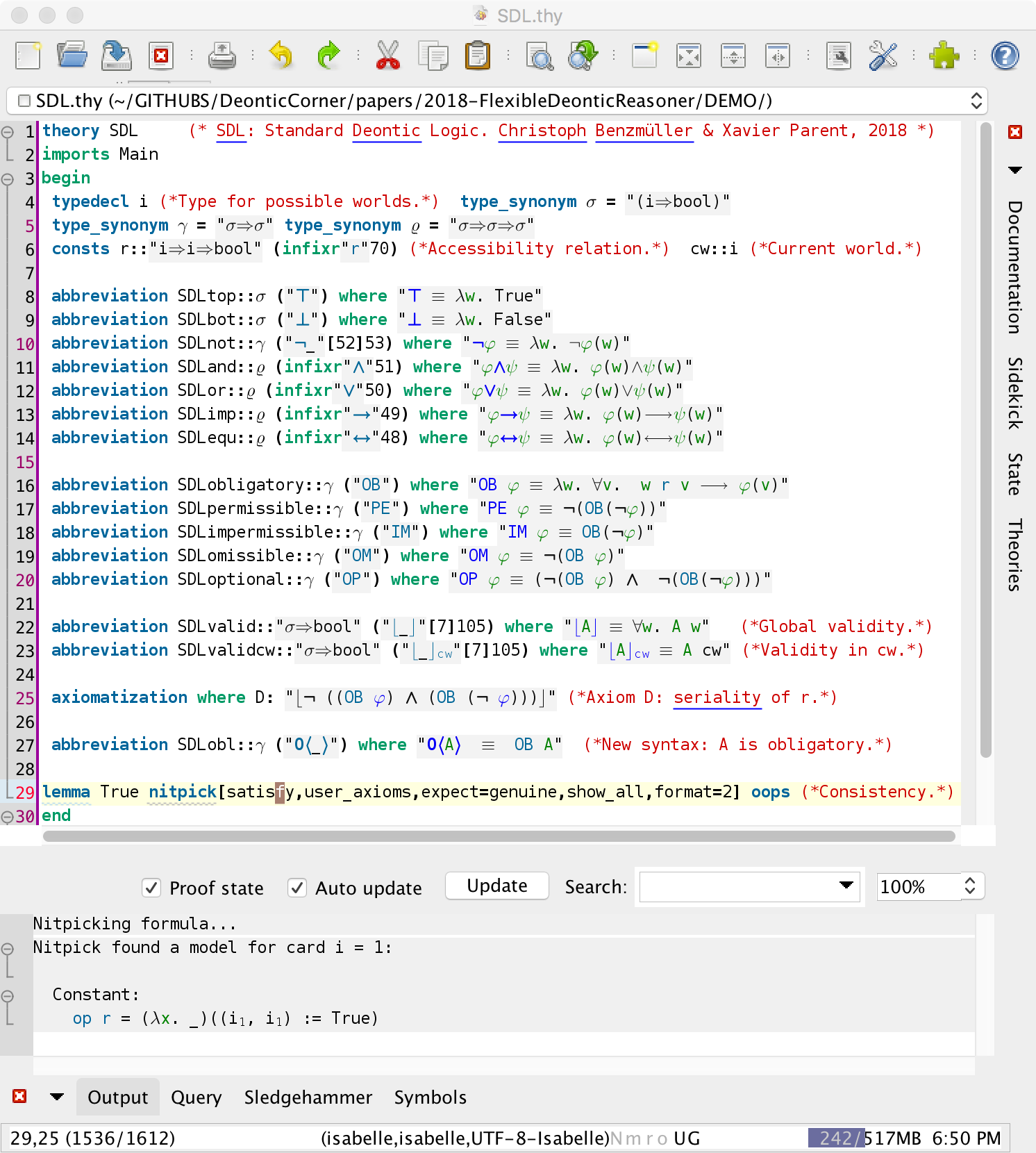}
\caption{Standard Deontic Logic in HOL. \label{fig1}}
\end{figure}

\subsubsection*{Standard Deontic Logic.}
Figure \ref{fig1} presents a shallow semantical embedding of SDL in Isabelle/HOL. SDL is synonymous for standard modal logic D, whose Kripke style semantics assumes a serial accessibility  relation. 
The core idea os this embedding is to lift SDL propositions to predicates on worlds. The world dependency of SDL propositions is thus  made explicit here, and the connectives are lifted accordingly. In other words, what is presented here is an encoding of the well known standard translation as a lean and elegant set of equations in Isabelle/HOL. 
 This semantical embedding of propositional SDL can easily be extended to first-order and even higher-order SDL, cf.~\cite{C62}.

The SDL unary obligation operator $\textbf{OB}$ is defined in line 16, and a standard syntax $\textbf{O}\langle.\rangle$ for unary obligation is then 
introduced in line 27. $\textbf{O}\langle \varphi \rangle$ is defined for the moment as $\textbf{OB} \varphi$. The idea is that $\textbf{O}\langle.\rangle$ can later be overloaded with an alternative notion of unary obligation defined in DDL.  The motivation for this overloading approach will become more transparent in Section \ref{sec:experiments}, where we switch between the two alternative definitions of  $\textbf{O}\langle \varphi \rangle$ by alternating the theory import in Isabelle/HOL only, while avoiding any changes in the actual formalisation of our small example.

\begin{figure}[tp] \centering
\includegraphics[width=\textwidth,height=.65\textheight]{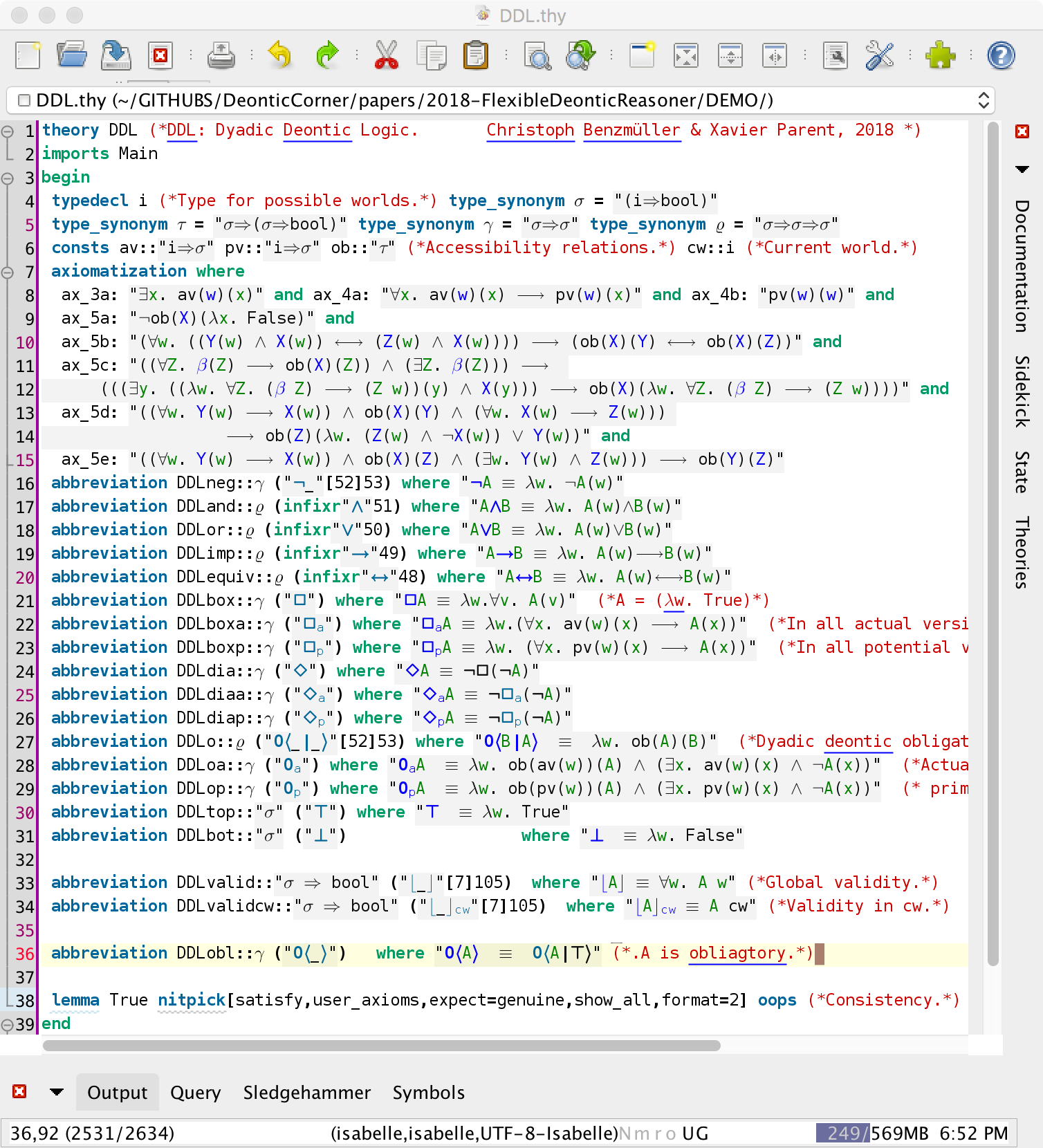}
\caption{Dyadic Deontic Logic in HOL. \label{fig2}}
\end{figure} 

\subsubsection*{Dyadic Deontic Logic.}
Dyadic deontic logic is the logic for reasoning with dyadic obligations (``it ought to be the case that \ldots if it is the case that \ldots''). 
Figure \ref{fig2} presents a shallow semantical embedding of a Dyadic Deontic Logic (DDL) proposed by Carmo and Jones \cite{CJ13,Carmo2002}
in Isabelle/HOL. Instead of a Kripke style semantics, DDL employs a neighborhood semantics (cf.~\cite{chellas1980modal}). This explains the change in the behaviour of the GDPR example below. In contrast to SDL, the DDL logic of Carmo and Jones is known to be robust against contrary-to-duty scenarios. This aspect will be further explored in Section \ref{sec:experiments}.

An unary obligation operator in DDL can be defined as a special case of  dyadic obligation where the condition  clause is set to $True$. In the Isabelle/HOL embedding of Fig.~\ref{fig2} this is done in line 36, where $\textbf{O}\langle \varphi \rangle$ is defined as $\textbf{O}\langle \varphi \mid T \rangle$. For illustration purposes we will work in Section \ref{sec:experiments} first with this unary obligation operator only in order to practically demonstrate the very different behaviour of it in a CTD scenario depending on whether we use SDL or DDL as the deontic logic of choice.

\section{Example: A CTD Situation in the GDPR} \label{sec:experiments}

For small, practical experiments in this section the General Data Protection Regulation (GDPR, Regulation EU
2016/679) 
has been chosen as an application scenario.
It is a regulation by which the European Parliament, the Council of the European Union and the European Commission intend to strengthen and unify data protection for all individuals within the European Union (EU). It also addresses the export of personal data outside the EU. The GDPR aims primarily to give control back to citizens and residents over their personal data and to simplify the regulatory environment for international business by unifying the regulation within the EU. The regulation was adopted on 27 April 2016. It becomes enforceable from 25 May 2018.  




We present below two sample norms contained in this knowledge base.  

\vskip.6em
\noindent\fcolorbox{black}{green!20}{\begin{minipage}{.98\columnwidth} 
\begin{enumerate}
\item Personal data shall be processed  lawfully (Art. 5). For example, the data subject must have given consent to the processing of his or her personal data for one or more specific purposes (Art. 6/1.a). 
\item  If the personal data have been processed unlawfully (none of the requirements for a lawful processing applies), the controller has the obligation to erase the personal data in question without delay (Art. 17.d, right to be forgotten).
\end{enumerate}
\end{minipage}}
\vskip.6em

When being combined with the following two knowledge units, the above GDPR norms exhibit a typical CTD-structure. 
\vskip.6em
\noindent
\fcolorbox{black}{green!20}{\begin{minipage}{.98\columnwidth} 
\begin{enumerate} \setcounter{enumi}{2}
\item \label{aa} It is an obligation (e.g. as part of a respective agreement between a customer and a company) to keep the personal data (as relevant to the agreement) provided that it is processed lawfully. 
\item \label{bb} I Some data in the context of such an agreement has been processed unlawfully.
\end{enumerate}
\end{minipage}}
\vskip.6em
The latter information pieces are not explicit parts of the GDPR. Instead they are to be seen as implicit.
\ref{aa} comes from a another regulation, with which the GDPR has to co-exists. \ref{bb} is a factual information --- it is exactly the kind of world situations the GDPR wants to regulate. 

To enable automated reasoning about GDPR enforced obligations (or permissions) a given world context the regulatory content of the GDPR will thus have to converge with a respective representation of this given situation. This can be seen as analogous to the notions of a T-Box (terminology) and an A-Box (world assertions) in semantic web applications.

In the remainder of this paper we will demonstrate, within a practical theorem proving environment, how critical the ``right'' choice of the assumed deontic logic actually is. While the illustrated effects are  known and well studied in the deontic logic literature, they are here for the first time exhibited and explored within an implemented, flexible deontic logic reasoner. Note in particular that DDL has never been implemented before. 


\subsubsection*{GDPR Example in SDL.}
\begin{figure}[tp] \centering
\includegraphics[width=\textwidth,height=.65\textheight]{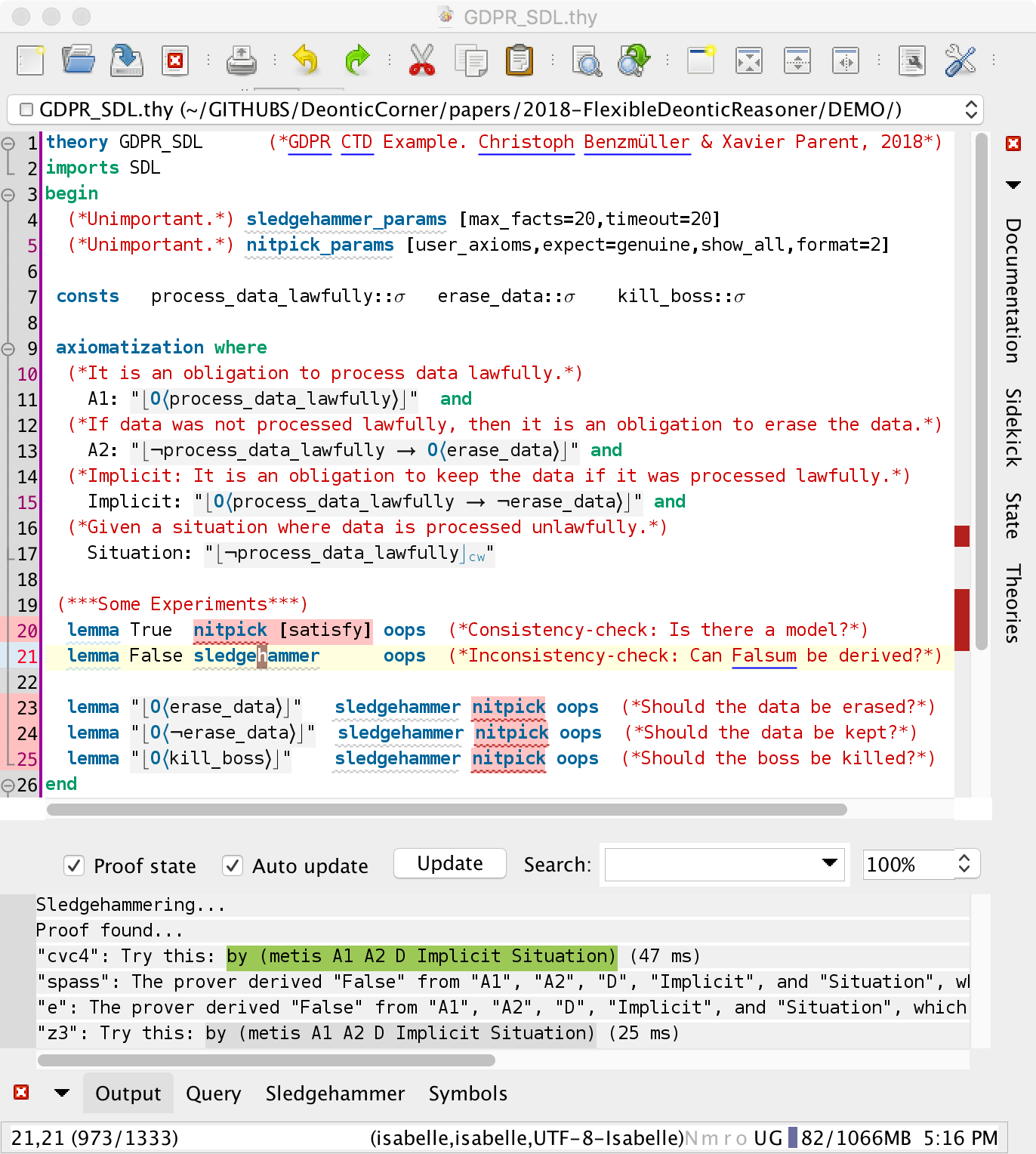}
\caption{GDPR related CTD example scenario in SDL. \label{fig:SDL}}
\end{figure}

Figure \ref{fig:SDL} presents the modeling of the scenario as discussed above in Isabelle/HOL.  In line 2 the deontic logic of choice is imported. For the moment this is SDL. 
The effect of the import is that the definitions of the semantical embedding of SDL in HOL as presented in Fig.~\ref{fig1} are loaded and activated.
In other words, the logical connectives from Fig.~\ref{fig1}, including the unary obligation operator $\textbf{O}\langle \varphi \rangle$,  are now available for the modeling of the example scenario.

In line 7 uninterpreted constant symbols are introduced. \textit{process\_data\_lawfully} encodes whether or not data has been processed lawfully in a given situation. \textit{erase\_data} encodes whether or not the data should be erased in the given situation. \textit{kill\_boss} has been added to
highlight the potential danger of  CTD scenarios in a particularly dramatic way. It shall encode the supposedly unrelated question whether the boss (e.g. of a company)  should be killed, which obviously must not be inferable as an obligation from the knowledge base.  Any other unrelated question would do equally well. 

It is relevant to remark that the propositional encoding as presented here is abstracting away lots of relevant structural information. For example, with data we rather mean something like ``data related to a particular person in the context of given costumer agreement''. To achieve a proper modeling of the entire GDPR a structurally more fine-grained content encoding in first-order or even higher-order deontic logic may thus be more adequate.

The previously mentioned GDPR norms are encoded in lines 10-13. In lines 14-17, the implicit knowlede, respectively the given world situation, is modeled.

Then, in lines 20-25, some experiments with the automated reasoning tools integrated with Isabelle/HOL are conducted.  In line 20 an initial check for consistency (satisfiability of the user axioms) with the model finder Nitpick \cite{Nitpick} fails.  This means the displayed axioms in lines 10-17 are inconsistent when SDL is taken as the deontic logic of choice. Moreover, this inconsistency is confirmed by the automated theorem provers (ATPs) CVC4 \cite{CVC4}, Spass \cite{Spass}, E \cite{E} and Z3 \cite{Z3}, which show that Falsum is inferred by the axioms (line 21).  These state-of-the-art ATPs are integrated with Isabelle/HOL  via the Sledgehammer \cite{Sledgehammer} tool.

The consequences of an inconsistency in a classical logic context are known --- everything follows (principle of explosion). This becomes apparent in our experiments in lines 23-25. In line 23, the ATPs prove that it is an obligation to erase the data. But in line 24 they also prove that the data should not erased. And most dramatically, in line 25, the ATPs prove that there is now the obligation to kill the boss. All these results ar confirmed also by the model finder Nitpick, which is employed here in countermodel finding mode to search for counterarguments to the statements in question. The red markup color indicates that Nitpick fails to do so.

From our first experiments we thus see that an SDL based deontic logic reasoner may easily turn into a dangerously irrational entity when being exposed to CTD scenarios. In oter words, the coice of the right deontic logic critically matters, and SDL is not a good choice.

\begin{figure}[tp] \centering
\includegraphics[width=\textwidth,height=.65\textheight]{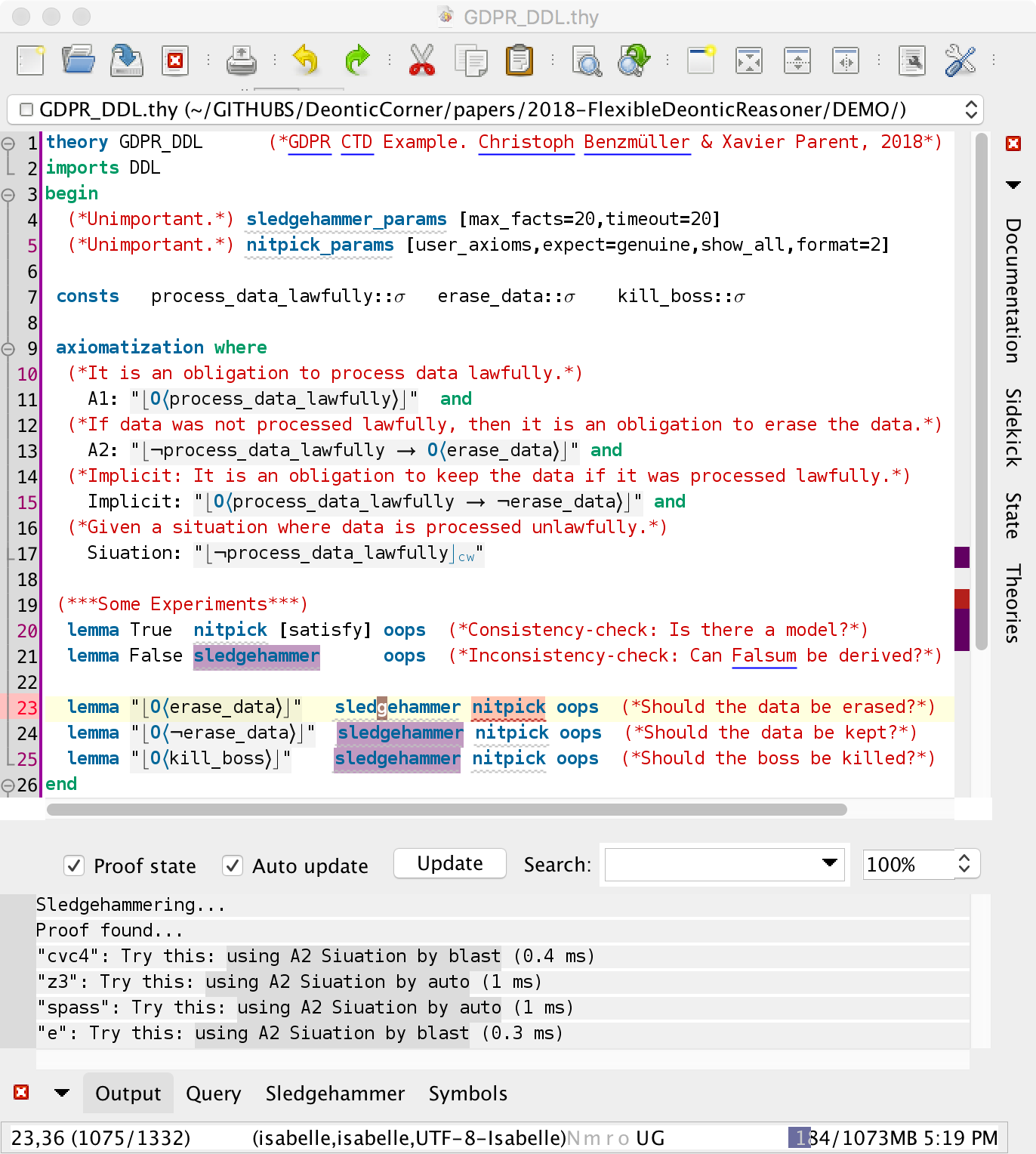}
\caption{GDPR related CTD example scenario in SDL. \label{fig:DDL}}
\end{figure}

\subsubsection*{GDPR Example in DDL (using unary obligation only).}
The very same experiments from Fig.~\ref{fig:SDL} are now repeated in Fig.~\ref{fig:DDL}. This time, however, DDL is selected and activated as the deontic logic of choice. While the modeling of the norms of the GDPR and the given situation remain exactly as before (lines 10-17), the experiment results are very different now. This time our reasoning infrastructure responds with the desired behaviour, due to choice of a more adequate deontic logic.

In line 20, the consistency check with Nitpick succeeds (no red markup), and in line 21 the attempts to prove Falsum with the ATPs fails.  In line 23, the ATPs prove that is in obligation in the given situation to erase the data, and Nitpick finds no counterargument to this.  These results are in-line with the negative response of the ATPs to the question whether the data shall be kept (line 24). Nitpick, in fact, presents a counterargument to this query.  And similarly the reasoning tools reject the query whether the boss should be killed.


\section{Conclusion}
Contrary-to-duty scenarios may arise in the context of recent regulatory frameworks such as the GDPR. However, it may difficult to detect them, and it may be even more difficult to properly address them. A particular challenge in this context is the choice of a suitable deontic logic formalism. The notion of suitability thereby at least includes robustness against contrary-to-duty scenarios. Several further requirements must be met, ranging from practical considerations, such as effective proof automation, to yet undetected, further theoretical challenges.   We therefore argue for the development of a flexible normative reasoning infrastructure enabling empirical studies with different deontic logic formalisms in the context of concrete application studies in which regulatory frameworks such as the GDPR are formalised and rigorously assessed. A starting point for the development of such a framework has been presented in this paper.

\bibliographystyle{abbrv}

\end{document}